\documentclass{article}
\usepackage{spconf,amsmath,graphicx}
\usepackage{algpseudocode}
\usepackage{algorithm}

\usepackage{multirow}
\usepackage{multicol}

\title{A COMPARISON OF LATTICE-FREE DISCRIMINATIVE TRAINING CRITERIA FOR PURELY SEQUENCE-TRAINED NEURAL NETWORK ACOUSTIC MODELS}
\name{Chao Weng, Dong Yu}
\address{Tencent AI Lab, Bellevue, USA \\
\tt \small  \{cweng,dyu\}@tencent.com}
\begin{document}
%
\maketitle
\begin{abstract}

In this work, three lattice-free (LF) discriminative training criteria for purely sequence-trained neural network acoustic models are compared on LVCSR tasks, namely maximum mutual information (MMI), boosted maximum mutual information (bMMI) and state-level minimum Bayes risk (sMBR). We demonstrate that, analogous to LF-MMI, a neural network acoustic model can also be trained from scratch using LF-bMMI or LF-sMBR criteria respectively without the need of cross-entropy pre-training. Furthermore, experimental results on Switchboard-300hrs and Switchboard+Fisher-2100hrs datasets show that models trained with LF-bMMI consistently outperform those trained with plain LF-MMI and achieve a relative word error rate (WER) reduction of $\sim$5\% over competitive temporal convolution projected LSTM (TDNN-LSTMP) LF-MMI baselines.             
 
\end{abstract}
\section{Introduction}
\label{sec:intro}

Sequence discriminative training has been proven to improve neural network (NN) acoustic models (AMs) significantly for LVCSR tasks \cite{lattice-based-mmi,dnn_smbr,seq-train-ms,asgd-smbr,embr}. Most widely used discriminative training criteria include maximum mutual information (MMI) \cite{ibm-mmi,largescale-mmi}, minimum classification error (MCE) \cite{mce}, boosted MMI (bMMI) \cite{bmmi}, minimum phone/word error (MPE/MWE) \cite{danthesis} and state-level minimum Bayes risk (sMBR) \cite{dnn_smbr,asgd-smbr,framelevelMPE}. Discriminative training of a NN is typically initialized from a frame-level cross-entropy pre-trained model which is also used to generate alignments and lattices, i.e, the reference and hypothesis sequences. Recently, lattice-free (LF) MMI \cite{lfmmi} was proposed to train a NN from scratch using MMI criterion without initializing from a cross-entropy model and achieved state-of-the-art performances on various LVCSR tasks. To make LF-MMI training practically feasible, a phone-level, instead of word-level, language model (LM) is used to generate a unified denominator cyclic graph, eliminating the need of the lattice generation process. LF-MMI has gained its popularity lately as it not only saves the cross-entropy pre-training effort but also leads to a better sequence-trained model in general for LVCSR. LF-MMI and LF-bMMI were adopted in \cite{mshuman} and \cite{multi-talker-bmmi} respectively on English conversation transcription and multi-talker speech recognition task but in both work cross-entropy pre-training is still required. Later on, various model architectures \cite{tdnn-lstm,self-attention-lfmmi,OPGRU-lfmmi} for LF-MMI have been explored. LF-sMBR was proposed and compared against lattice-based sMBR in \cite{lf-smbr} when initialized from a LF-MMI trained model. Very recently, various lattice-free discriminative training criteria are investigated for keyword spotting task \cite{dt-kws}. However, there are few, if any, prior works having presented a comparison between different discriminative training criteria with which NN AMs are purely sequence-trained from scratch for LVCSR tasks.

In this work, three lattice-free discriminative training criteria for purely sequence-trained NN AMs are compared on LVCSR tasks, MMI, bMMI and sMBR. We demonstrate that, analogous to LF-MMI, a NN AM can also be trained from scratch using LF-bMMI or LF-sMBR criteria respectively without the need of cross-entropy pre-training. Furthermore, experimental results on Switchboard-300hrs and Switchboard+Fisher-2100hrs datasets show that models trained with LF-bMMI consistently outperform those trained with plain LF-MMI and achieve a relative word error rate (WER) improvement of $\sim$5\% over competitive TDNN-LSTMP LF-MMI baselines. 

The remainder of the paper is organized as follows. In Section 2, we elaborate lattice-free discriminative training criteria and then describe implementation details of LF-bMMI and LF-sMBR. All the experimental setups and results are presented in Section 3. We conclude our work in Section 4.

\section{Lattice-free Sequence Discriminative Training}
\label{sec:lfsdt}
\subsection{LF-MMI}
\label{ssec:lfmmi}
Given $U$ pairs of the training speech utterance $\mathbf{O}_u$  and its reference word transcription $\mathbf{W}_u^r$, the objective function of LF-MMI is given by, 
\begin{align}
\mathcal{F}_{\mathrm{LF-MMI}} = \sum_{u=1}^{U}\log \frac{\sum_{\mathbf{S}_u^r} p(\mathbf{O}_u | \mathbf{S}_u^r) P (\mathbf{W}_u^r)}{\sum_{\mathbf{S}_u} p(\mathbf{O}_u | \mathbf{S}_u) P (\mathbf{W}_u) },
\label{eq:lfmmiobj}
\end{align}
where $\mathbf{W}_u$ is a hypothesis sequence from the denominator graph. $\mathbf{S}_u^r$ and $\mathbf{S}_u$ are state sequences corresponding to $\mathbf{W}_u^r$ and $\mathbf{W}_u$ respectively. Note that there could be multiple $\mathbf{S}_u$ corresponding to one $\mathbf{W}_u$ due to multiple pronunciations and label tolerance, i.e., a tolerance which allows a state to occur slightly before or after where it appear in the original reference alignments or lattices \cite{lfmmi}. Let $\mathbf{o}_t$ and $\mathbf{s}_t$ be the $t$th frame of $\mathbf{O}_u$ and $t$th state of $\mathbf{S}_u$ respectively, i.e., $\mathbf{O}_u = \{\mathbf{o}_1,...,\mathbf{o}_t,...,\mathbf{o}_{T_u}\}$, $\mathbf{S}_u = \{\mathbf{s}_1,...,\mathbf{s}_t,...,\mathbf{s}_{T_u}\}$. The NN output at time $t$ is interpreted as pseudo log-likelihood of $\mathbf{o}_t$ given a particular state $j$, i.e., $\log p(\mathbf{o}_t |\mathbf{s}_t = j )$, and there is neither acoustic scaling or division by the prior in LF-MMI. The gradient required to perform LF-MMI training can be written as,
\begin{align}
\frac{\partial \mathcal{F}_{\mathrm{LF-MMI}}}{ \partial \log p(\mathbf{o}_t |\mathbf{s}_t = j )} = \sum_{u=1}^U [\gamma_u^{\mathrm{num}} (\mathbf{s}_t = j) - \gamma_u^{\mathrm{den}} (\mathbf{s}_t = j)],
\label{eq:lfmmigrad}
\end{align}
where $\gamma_u^{\mathrm{num}} (\mathbf{s}_t = j)$ and $\gamma_u^{\mathrm{den}} (\mathbf{s}_t = j)$ are posterior probabilities of $t$th state being $j$ computed over the reference and hypothesis state sequences respectively given the input utterance $\mathbf{O}_u$.

\subsection{LF-bMMI}
\label{ssec:lfbmmi}
In LF-bMMI, the objective function is modified to boost the likelihoods of those hypothesis sequences commit more errors, 
\begin{align}
\mathcal{F}_{\mathrm{LF-bMMI}} = \sum_{u=1}^{U}\log \frac{\sum_{\mathbf{S}_u^r} p(\mathbf{O}_u | \mathbf{S}_u^r) P (\mathbf{W}_u^r)}{\sum_{\mathbf{S}_u} p(\mathbf{O}_u | \mathbf{S}_u) P (\mathbf{W}_u) e^{-b A(\mathbf{S}_u, \mathbf{S}_u^r)} },
\label{eq:lfbmmiobj}    
\end{align}
where $b$ is the boosting factor and $A(\mathbf{S}_u^r, \mathbf{S}_u)$ is the accuracy function which measures the number of matching labels between reference and hypothesis sequences. The gradient of LF-bMMI has the same form as LF-MMI in Eq. (\ref{eq:lfmmigrad}) whereas the difference from LF-MMI lies in the forward-backward implementation when calculating the $\gamma_u^{\mathrm{den}} (\mathbf{s}_t = j)$ term. See more details in Subsection \ref{ssec:impl}. 
\subsection{LF-sMBR}
The objective function of LF-sMBR criterion is,
\begin{align}
\mathcal{F}_{\mathrm{LF-sMBR}}= \sum_{u=1}^{U}  \frac{\sum_{\mathbf{S}_u} p(\mathbf{O}_u | \mathbf{S}_u) P (\mathbf{W}_u) A (\mathbf{S}_u, \mathbf{S}^r_u)}{\sum_{\mathbf{S}'_u} p(\mathbf{O}_u | \mathbf{S}'_u) P (\mathbf{W}'_u) }.
\end{align}
The gradient required to perform LF-sMBR training can be written as,
\begin{align}
\frac{\partial \mathcal{F}_{\mathrm{LF-sMBR}}}{ \partial \log p(\mathbf{o}_t |\mathbf{s}_t = j )} = \sum_{u=1}^U \gamma_u^{\mathrm{den}}(\mathbf{s}_t = j)  [\overline{A}_u(\mathbf{s}_t = j) - \overline{A}_u ],
\label{eq:lfsmbrobj}    
\end{align}
where $\overline{A}_u(\mathbf{s}_t = j)$ is the average state-level accuracy of those hypothesis state sequences that have state $j$ at frame $t$ and $\overline{A}_u$ is the average accuracy of all state sequences in the denominator graph. $\gamma_u^{\mathrm{den}}(\mathbf{s}_t = j)$ is the same posterior probability term as in Eq. (\ref{eq:lfmmigrad}). 

\begin{algorithm}
\caption{LF-sMBR Forward Procedure for Leaky HMM}
\label{alg:forward}
\renewcommand{\algorithmicrequire}{\textbf{}}
\renewcommand{\algorithmicensure}{\textbf{Output:}}
\begin{algorithmic}[1]
\Require $T \leftarrow$ number of frames; $S \leftarrow$ number of HMM states in the denominator graph; $\lambda \leftarrow $ leaky HMM coefficient
\For{$s \leftarrow$ $1$ to $S$}
\State $P_0(s) \leftarrow$ initial probability of state $s$
\State $\alpha (s, 0) \leftarrow P_0(s)$ 
\State $\alpha_{\mathrm{mbr}} (s, 0) \leftarrow 0$ 
\EndFor
\For{$s \leftarrow$ $1$ to $S$}
\State $\alpha (s, 0) \leftarrow \alpha (s, 0) + \lambda P_0(s) \sum_1^S \alpha (s, 0) $ 
\State $\alpha_{\mathrm{mbr}} (s, 0) \leftarrow \alpha_{\mathrm{mbr}} (s, 0) + \lambda P_0(s) \sum_1^S \alpha_{\mathrm{mbr}} (s, 0)$ 
\State $\alpha_{\mathrm{mbr}} (s,0) \leftarrow \alpha_{\mathrm{mbr}} (s,0) /  \alpha (s, 0) $
\EndFor

\For{$t \leftarrow$ $1$ to $T$}
  \For {$s \leftarrow$ $1$ to $S$}
    \State $\alpha (s, t) \leftarrow 0$
    \State $\alpha_{\mathrm{mbr}} (s, t) \leftarrow 0$
    \For { arc $ a \in$ all arcs preceding state $s$}
    \State $P_{\mathrm{trans}} \leftarrow$ transition probability of arc $a$
    \State $j  \leftarrow $ state label (index of NN output) of arc $a$
    \State $s_{\mathrm{prev}}  \leftarrow $ starting state of arc $a$
    \State $\delta = \alpha (s_{\mathrm{prev}}, t-1) * P_{\mathrm{trans}} * p(\mathbf{o}_t |\mathbf{s}_t = j )$
    \State $\alpha (s, t) \leftarrow \alpha (s, t) + \delta$
    \State $\alpha_{\mathrm{mbr}} (s, t) \leftarrow [\alpha_{\mathrm{mbr}} (s_{\mathrm{prev}}, t-1) + \gamma_u^{\mathrm{num}} (\mathbf{s}_t = j)] * \delta $
    \EndFor 
  \EndFor
  \For {$s \leftarrow$ $1$ to $S$}
  \State $\alpha (s, t) \leftarrow \alpha (s, t) + \lambda P_0(s) \sum_1^S \alpha (s, t) $ 
  \State $\alpha_{\mathrm{mbr}} (s, t) \leftarrow \alpha_{\mathrm{mbr}} (s, t) + \lambda P_0(s) \sum_1^S \alpha_{\mathrm{mbr}} (s, t)$ 
  \State $\alpha_{\mathrm{mbr}} (s,t) \leftarrow \alpha_{\mathrm{mbr}} (s,t) /  \alpha (s, t) $
  \EndFor
\EndFor
\end{algorithmic}
\end{algorithm}

\subsection{Implementation of LF-bMMI and LF-sMBR}
\label{ssec:impl}
Since LF-MMI implementation is available in the Kaldi toolkit \cite{kaldi}, we mainly discuss the implementation details of LF-bMMI and LF-sMBR in this subsection. Due to multiple pronunciations and label tolerance, the supervision used in LF-MMI is a numerator graph which encodes multiple reference state sequences. In both our LF-bMMI and LF-sMBR implementations, the term $\gamma_u^{\mathrm{num}} (\mathbf{s}_t = j)$ derived from the numerator graph is used to calculate per-frame state-level accuracy. 

For LF-bMMI, the only extra computation as opposed to plain LF-MMI is to boost per-frame likelihood according to $\gamma_u^{\mathrm{num}} (\mathbf{s}_t = j)$ when forward-backward is performed on the denominator graph, specifically,
\begin{align}
p'(\mathbf{o}_t |\mathbf{s}_t = j ) = p(\mathbf{o}_t |\mathbf{s}_t = j ) \times e^{-b \times \gamma_u^{\mathrm{num}} (\mathbf{s}_t = j) }.
\label{eq:boostlikelihood}
\end{align}

 For LF-sMBR, the implementation closely follows lattice-based sMBR but there are two notable variations. The first one lies in the leaky hidden Markov model (HMM) regularization \cite{lfmmi} used in LF-MMI where an $\epsilon$ transition from each state $a$ to each other state $b$ is added to the denominator HMM graph with probability equal to a coefficient times the initial probability of state $b$. Since the leaky HMM is enabled in both our LF-MMI and LF-bMMI experiments, we also incorporate it in our LF-sMBR implementation. We list the implementation details of LF-sMBR forward procedure on a denominator leaky HMM graph in Algorithm \ref{alg:forward} where Line 7, 8, 25 and 26 correspond to the forward computations for leaked $\epsilon$ transitions. Note that per-frame state-level accuracy for all $\epsilon$ transitions is zero. The second variation is we only need one single forward-backward pass during LF-sMBR training for efficiency as opposed to the lattice based sMBR implementation in the Kaldi \cite{kaldi} which requires two forward-backward passes. The backward procedure can be easily derived given the forward procedure listed in Algorithm \ref{alg:forward}.

\section{Experiments}
\label{sec:exp}

We conduct our experiments on both Switchboard-300hrs and Switchboard+Fisher-2100hrs datasets to compare LF-MMI, LF-bMMI and LF-sMBR for LVCSR. Speed perturbation is used to augment the training data three times in both Switchboard-300hrs and Switchboard+Fisher-2100hrs experiments. We use the same set of numerator and denominator lattices/graphs for all three training criteria. The numerator graphs are generated using a GMM-HMM system trained on LDA+STC+FMLLR MFCC features described in \cite{dnn_smbr} and the denominator graph is generated using a phone language model estimated from the phone alignments on all training transcriptions using the GMM-HMM system. TDNN-LSTMP architecture \cite{tdnn-lstm,OPGRU-lfmmi} is adopted in all our experiments as it was found to outperform bidirectional LSTMs and achieve state-of-the-art performances on the Switchboard-300hrs dataset when trained with LF-MMI. We listed the detailed model configuration used in all of our experiments in Table \ref{tab:tdnn-lstm-arch}. The input features to the TDNN-LSTMP are 40-dim MFCCs and 100-dim online i-vectors. All TDNN-LSTMPs are trained from scratch without any type of pre-training but we apply cross-entropy regularization during training and the smooth factor is set to 0.025 unless otherwise indicated. The leaky HMM coefficient is set to 0.1 through all our experiments. The initial and final learning rates are set to 0.001 and 0.0001 respectively for all three criteria. We adopt the default parallel optimizer \cite{natural-gradient-average} in the Kaldi toolkit \cite{kaldi} for all training experiment and use 3 GPUs initially and ramp up to 8 in the end to train TDNN-LSTMPs. Note that there is no early stop for this optimizer, one has to specify a fixed number of training epochs before each experiment. Both tri-gram and 4-gram LMs are estimated on Switchboard+Fisher-2100hrs transcriptions and all WER results are reported using the 4-gram LM to re-score the lattices that are generated with the trigram LM.
\begin{table}
\centering
\caption{The model configuration used in all our experiments}
\vspace{0.3cm}
\begin{tabular}{c|cc}
\hline 
\hline
Layer & Context & Layer-type \\
\hline
1 & [-2,-1,0,1,2] & TDNN \\
2 & [-1,0,1] & TDNN \\
3 & [-1,0,1] & TDNN \\
4 & [0] & LSTMP \\
5 & [-3, 0, 3] & TDNN \\
6 & [-3, 0, 3] &TDNN \\
7 & [0] & LSTMP \\
8 & [-3, 0, 3] & TDNN \\
9 & [-3, 0 ,3] & TDNN \\
10 & [0] & LSTMP \\
11 & [0] & Projection \\
\hline
\end{tabular}
\label{tab:tdnn-lstm-arch}
\end{table}
\subsection{Experiments on Switchboard-300hrs}
After building a tree using the special two-state HMM topology described in \cite{lfmmi}, the output dimension of TDNN-LSTMPs on Switchboard-300hrs is 5994 which leads the number of TDNN-LSTMP's total parameters to 39.5 million. This is a pretty large model on 300hrs training data. Therefore, we use dropout and follow the same schedule in \cite{dropout-lfmmi} to overcome over-fitting. We first compare TDNN-LSTMP LF-MMI baseline with LF-bMMI using different boosting factors $b$ from 0.05 to 0.30 in Table \ref{tab:lfmmi-300hrs}. The number of training epochs for both LF-MMI and LF-bMMI is set to 4, which appears to be optimal according to the Kaldi toolkit \cite{kaldi}. It shows that TDNN-LSTMPs trained with LF-bMMI outperform the LF-MMI baseline consistently and the best LF-bMMI model achieves an absolute WER improvement of 0.4\% and 0.6\% on the Switchboard subset and full set of Eval2000 respectively.  
\begin{table}
\centering
\caption{WERs of TDNN-LSTMP LF-MMI baseline and LF-bMMI on Eval2000 using Switchboard-300hrs data}
\vspace{0.3cm}
\begin{tabular}{c|c|c|c}
\hline 
\hline
\multirow{2}{*}{Models} & \multirow{2}{*}{$b$} & \multicolumn{2}{c}{WERs(\%)}  \\
\cline{3-4}
  &  &  SWB & Total  \\
\hline
\hline
LF-MMI & 0.0  & 9.1 & 14.2 \\
LF-bMMI & 0.05 & 8.7 & 13.7 \\
LF-bMMI & 0.15 & 8.7 & 13.7 \\
LF-bMMI & 0.20& \textbf{8.7} & \textbf{13.6} \\
LF-bMMI & 0.25 & 8.8 & 13.6 \\
LF-bMMI & 0.30 & 8.8 & 13.7 \\
\hline
\end{tabular}
\label{tab:lfmmi-300hrs}
\vspace{-0.5cm}
\end{table}
Then LF-sMBR is compared against LF-MMI in Table \ref{tab:lfsmbr-300hrs}. During LF-sMBR experiments, it is found that if we use the original $\gamma_u^{\mathrm{num}} (\mathbf{s}_t = j)$ term for the silence state accuracy, the model diverges after a few tens of training iterations. So we simply apply a scale, eg., $0 < \mu < 1$, to $\gamma_u^{\mathrm{num}} (\mathbf{s}_t = j)$ if state $j$ belongs to silence. Note that we also tried zeroing the gradients for silence states, i.e., $\gamma_u^{\mathrm{den}} (\mathbf{s}_t = j) = 0 $ if $j\in \mathrm{silence}$, as in \cite{lf-smbr}, but the number of insertion errors is disproportionately high due to the fact that the silence model is not being trained well. Another critical configuration for LF-sMBR training is the label tolerance, which allows a state drifting away from where it appear in the original alignment. As shown in Table \ref{tab:lfsmbr-300hrs}, the setup using the numerator graph derived from the original alignments, i.e., no label tolerance, achieves better performances. We experiment different silence state accuracy scales from 0.001 to 0.02 and find small scales like 0.001 will lead to too many insertion errors due to insufficient silence modeling and a scale of 0.02 will make the training diverged. It is also found that LF-sMBR converges slower than LF-MMI or LF-bMMI, and there are improvements when we increase the number of training epochs from 8 to 12. The best LF-sMBR setup achieves WERs of 11.2\% and 15.8\% on the Switchboard subset and full set of Eval2000 which unfortunately still shows an inferior performance to the LF-MMI baseline. Therefore, we will only compare LF-MMI and LF-bMMI on the full Switchboard+Fisher-2100hrs dataset.
\begin{table}
\centering
\setlength{\tabcolsep}{3pt}
\caption{WERs of TDNN-LSTMP LF-MMI baseline and LF-sMBR with different configurations on Eval2000 using Switchboard-300hrs data}
\vspace{0.3cm}
\begin{tabular}{c|c|c|c|c|c}
\hline 
\hline
\multirow{2}{*}{\shortstack[c]{Models - \\  \#Epochs}} &\multirow{2}{*}{\shortstack[c]{Label \\ tolerance}} 
&\multirow{2}{*}{\shortstack[c]{x-ent \\smooth }} &
\multirow{2}{*}{\shortstack[c]{Sil-acc.\\ Scale}}  & \multicolumn{2}{c}{WERs(\%)}  \\
\cline{5-6}
& &  &  &  SWB & Total  \\
\hline
\hline
LF-MMI - 4 & $\pm$ 5 & 0.025 & NA & 9.1 & 14.2 \\
LF-sMBR - 4 & $\pm$ 5& 0.025 & 0.001 & 20.3 & 23.9 \\
LF-sMBR - 5 & $\pm$ 1 & 0.025 & 0.001 & 17.1 & 20.9 \\
LF-sMBR - 8 & $\pm$ 1 &  0.025& 0.001& 16.2 & 19.7 \\
LF-sMBR - 8 & $\pm$ 0& 0.025& 0.01 &  12.0 & 16.9 \\
LF-sMBR - 12 & $\pm$ 0& 0.025&0.01 & 11.6 & 16.2 \\
LF-sMBR - 12 & $\pm$ 0& 0.025&0.013 & \textbf{11.2} & \textbf{15.8} \\
LF-sMBR - 12 & $\pm$ 0& 0.025 &0.015 & 11.5 & 16.3\\
LF-sMBR - 12 & $\pm$ 0& 0.050 &0.013 & 11.2 & 16.0 \\
\hline
\end{tabular}
\label{tab:lfsmbr-300hrs}
\end{table}

\subsection{Experiments on Switchboard+Fisher-2100hrs}
We then use the full Switchboard+Fisher2000hrs dataset as training data to compare LF-MMI and LF-bMMI criteria. The output size of TDNN-LSTMP is 7266, a bit larger than the one used in the 300hrs experiment. As we have much larger amount of training data, the dropout is disabled for all 2100hrs experiments. For both LF-MMI and LF-bMMI experiments on 2100hrs, we set the number of training epochs to 4 and report the WER results in Table \ref{tab:lfmmi-2100hrs}. The LF-MMI baseline obtain WERs of 8.1\% and 15.5\% on Switchboard and Callhome subsets of Eval2000 which are very competitive to those results reported in \cite{tdnn-lstm} and \cite{dropout-lfmmi}. Three LF-bMMI with different boosting factors are explored and the best setup achieves WERs of 7.7\% and 14.7\% on Switchboard and Callhome subsets respectively, a $\sim$5\% relative WER improvement over the TDNN-LSTMP LF-MMI baseline.
\begin{table}
\centering
\caption{WERs of TDNN-LSTMP LF-MMI baseline and LF-bMMI on Eval2000 using Switchboard+Fisher-2100hrs data}
\vspace{0.3cm}
\begin{tabular}{c|c|c|c|c}
\hline 
\hline
\multirow{2}{*}{Models} & \multirow{2}{*}{$b$} & \multicolumn{3}{c}{WERs(\%)}  \\
\cline{3-5}
  &  &  SWB & CH & Total  \\
\hline
\hline
LF-MMI & 0.0  & 8.1 & 15.5 &  12.0 \\
LF-bMMI & 0.05 & 8.1 & 15.2 &  11.7\\
LF-bMMI & 0.10 & \textbf{7.7} & \textbf{14.7} &  \textbf{11.3} \\
LF-bMMI & 0.15&  7.9 & 14.9 & 11.5 \\
\hline
\end{tabular}
\label{tab:lfmmi-2100hrs}
\vspace{-0.3cm}
\end{table}
Finally we compare our best LF-bMMI trained TDNN-LSTMP with previous published systems. Note that we only consider the reported results using similar n-gram LMs. As can be seen in Table \ref{tab:systemcompare}, our best LF-bMMI trained TDNN-LSTMP is highly competitive to those state-of-the-art systems.  
\begin{table}[H]
\setlength{\tabcolsep}{3pt}
\caption{Comparing our best model to previous published systems built on the Switchboard+Fisher-2000hrs.}
\vspace{0.3cm}
\centering
\begin{tabular}{c|c|c}
\hline
\hline
\multirow{2}{*}{Systems}  & \multicolumn{2}{c}{WERs(\%)}  \\
\cline{2-3}
  &    SWB & CH  \\
\hline
\hline
  BLSTM+LFMMI  \cite{lfmmi} & 8.5 & 15.3 \\
  ResNet+xent+LFMMI \cite{mshuman} & 8.6 & 15.2 \\ 
  LACE+xent+LFMMI  \cite{mshuman} & 8.5 & 15.2 \\ 
  ResNet+xent+MMI  \cite{ibm2016} & 8.3 & 14.9 \\ 
  TDNN-NormOPGRU+LFMMI \cite{OPGRU-lfmmi} & 8.3 & 14.7 \\
  TDNN-LSTMP+LFbMMI (current) & 7.7 & 14.7 \\
\hline
\end{tabular}
\label{tab:systemcompare}
\end{table}

\section{Conclusion}
\label{sec:conclusion}

In this work, three lattice-free (LF) discriminative training criteria for purely sequence-trained NN AMs are compared on LVCSR tasks, MMI, bMMI and sMBR. We demonstrate that, analogous to LF-MMI, a NN AM can also be trained from scratch using LF-bMMI or LF-sMBR criteria respectively without the need of cross-entropy pre-training. Furthermore, experimental results on Switchboard-300hrs and Switchboard+Fisher-2100hrs datasets show that models trained with LF-bMMI consistently outperform those trained with plain LF-MMI and achieve a WER reduction of $\sim$5\% over competitive TDNN-LSTMP LF-MMI baselines. Future work includes investigating why LF-sMBR falls behind both LF-MMI and LF-bMMI significantly and combining LF-sMBR with LF-MMI for multi-task sequence-trained NN AMs.


\bibliographystyle{IEEEbib}
\bibliography{manuscript}

\end{document}